\documentclass{article}
\usepackage{amsmath}
\usepackage{amsfonts}
\usepackage{cite}
\usepackage{geometry}
\usepackage{booktabs}
\usepackage{multirow}
\usepackage{algorithm}
\usepackage{algpseudocode}
\usepackage{authblk}

\begin{document}

\title{A Framework for Non-Linear Attention via Modern Hopfield Networks}
\author{A. Farooq}
\date{} 

\affil[]{University of New Brunswick}
\parindent 0ex

\maketitle

\begin{abstract}
In this work we propose an energy functional along the lines of Modern Hopfield Networks (MNH), the stationary points of which correspond to the attention due to Vaswani et al. \cite{Vaswani2017}, thus unifying both frameworks.  The minima of this landscape form``context wells''—stable configurations that encapsulate the contextual relationships among tokens.  A compelling picture emerges: across $n$ token embeddings  an energy landscape is defined whose gradient corresponds to the attention computation.  Non-linear attention mechanisms offer a means to enhance the capabilities of transformer models for various sequence modeling tasks by improving the model's understanding of complex relationships, learning of representations, and overall efficiency and performance. A rough analogy can be seen via cubic splines which offer a richer representation of non-linear data where a simpler linear model may be inadequate. This approach can be used for the introduction of non-linear heads in transformer based models such as BERT, \cite{devlin2019bert}, etc. 
\end{abstract}

\section{Introduction}

The concept of energy serves as a cornerstone in the study and operation of Hopfield networks, which are recurrent neural networks designed to emulate associative memory. Originally proposed in 1982 by John Hopfield \cite{Hopfield1982}, these networks rely on an energy functional to guide their dynamics toward stable states that represent stored memories. The energy functional for the classic Hopfield network is expressed as:

\begin{equation}
E = -\frac{1}{2} \sum_{i=1}^N \sum_{j=1}^N w_{ij} s_i s_j
\end{equation}

where $s_i$ denotes the state of the $i$-th neuron, and $w_{ij}$ represents the synaptic weight between neurons $i$ and $j$. This energy function, which the network minimizes during its iterative updates, enables the system to converge to configurations that correspond to previously stored patterns, facilitating memory retrieval even from partial or corrupted inputs. The quadratic nature of this energy formulation underscores its role in defining the network’s stable states, a principle that has profoundly influenced subsequent developments in neural network theory.\\

Further advancing this framework, Amit et al. \cite{Amit1985}, explored the Hopfield model through the lens of statistical mechanics, enhancing its applicability by incorporating more realistic neuronal dynamics. Their analysis elucidated the network’s storage capacity and stability properties, particularly under conditions where the number of stored patterns approaches the network’s theoretical limits or when noise perturbs the system. By treating the network as a spin-glass system, they provided a deeper understanding of how energy minimization governs the retrieval process, laying the groundwork for more robust associative memory models.\\

In 2016 Krotov and Hopfield \cite{Krotov2016} introduced Modern Hopfield Networks (MHN) by introducing higher-order interactions among neurons through a generalized energy functional $F$ of the form:

\begin{equation}
E = \sum_{\mu=1}^K F\left( \sum_{i=1}^N \xi_i^\mu s_i \right)
\label{eqn:1a}
\end{equation}

where $\xi_i$ are the stored memories and $s_i$ is the state vector leading to higher memory storage capacities based on polynomial forms of $F$.    Drawing on this, Demircigil et al. \cite{demircigil2017model} proposed the form $F(s_i) = \exp (\xi^\mu_i s_i)$ and showed that it had exponential storage capacity.  Finally, Ramsauer et al. \cite{ramsauer2021hopfield} constructed an energy functional based on a log-sum-exp functional, and demostrated that this functional shares a deep connection with the attention mechanism introduced by Vaswani et al. [1]. Specifically, the fixed points of MHN dynamics align with the computational structure of attention. For a state vector $x_i$, a simplified energy functional in MHNs can be expressed as:

\begin{equation}
E(\mathbf{x},\beta) = \frac{1}{2} \mathbf{x}^T\mathbf{x} + \log \left( \sum_i \exp (\beta x_i) \right)
\end{equation}

where $\beta > 0$ is a constant controlling the sharpness of the exponential terms. 
In the general formulation, this involves stored patterns $\xi_i$, but under the assumption $\xi_i = e_i$ (standard basis vectors),  
the energy reflects interactions within $\mathbf{x}$. At equilibrium, where the gradient vanishes ($\nabla E_R = 0$) the state satisfies:

\begin{equation}
\mathbf{x} =  \beta \; \text{softmax} \left(\beta x_i \right)
\label{eqn:1}
\end{equation}

mirroring the softmax-based weighting in attention mechanisms, suggesting that MHNs compute a form of self-attention at their fixed points.\\

In another development, Hoover et al. \cite{Hoover2023} introduced a novel Transformer architecture based on the principle of minimizing a carefully designed energy function which differs from the standard stacked feedforward structure of Transformers, by employing a single, recurrent Energy Transformer (ET) block. with and energy function 

\begin{equation}
E = - \frac{1}{\beta} \sum_{C} \sum_{h} \log \left( \sum_{B } \exp \left( \beta \left( \sum_{\alpha} K_{ahB} Q_{ahC} \right) \right) \right) - \frac{1}{2} \sum_{C, \mu} G \left( \sum_{j} \xi_{\mu j} g_{jC} \right) 
\end{equation}

The update rule for the token representations $x_j$ within this block is governed by the gradient of the total energy function. In another study, Krotov and Hopfield, \cite{krotov2021large} have modeled neuronal dynamics as a set of coupled Lagrangians. \\

The Vaswani attention mechanism, Vaswani et al. \cite{Vaswani2017}, foundational to transformer models is given below:

\begin{equation}
\text{Attention}(Q, K, V) = \text{softmax}\left(\frac{QK^T}{\sqrt{d_k}}\right)V
\label{eqn:2}
\end{equation}

where  $Q, K \in \mathbb{R}^{n \times d_k}$ and $V \in \mathbb{R}^{n \times d_v}$ and $A \in \mathbb{R}^{n \times n}$ represent the query, keys, values and attention respectively. It can be seen that $1/d_k$, like $\beta$ in MHN's, scaling the softmax distribution similarly to an inverse temperature in Maxwell-Boltzmann statistics, which governs the focus of attention weights. This mechanism addresses a critical limitation of recurrent neural networks (RNNs) by enabling parallel processing of long sequences within a block structure, a capability long exploited in convolutional neural networks (CNNs). This parallelization enhances efficiency and scalability, yet the similarity of equations \ref{eqn:1} and \ref{eqn:2} prompts a question: is it possible to formulate an energy functional that casts the Vaswani attention mechanism as a gradient descent process?\\

\section{An Energy Functional for the Self-Attention Mechanism}

The Query matrix $Q \in \mathbb{R}^{n \times d_k}$, the key matrix $K \in \mathbb{R}^{n \times d_k}$ may be written as $Q = [q_{\text{1}}^T q_{\text{2}}^T \ldots q_{\text{n}}^T]^T$, $K = [ k_{\text{1}}^T k_{\text{2}}^T \ldots k_{\text{n}}^T]^T$, with $q_i, k_i \in \mathbb{R}^{d_k}$.  In index notation the argument of the softmax function (see equation \ref{eqn:2}) can be written as:

\[
\left( \frac{Q K^T}{\sqrt{d_k}} \right)_{ij} = \frac{q_i^T k_j}{\sqrt{d_k}}
\]

Using $\zeta_i = \sum_{m=1}^n \exp\left( \frac{q_i^T k_m}{\sqrt{d_k}} \right)$\footnote{In statistical mechanics the partition function is customarily denoted using $Z$, however we will use $\zeta_i$ since we will be using $Z$ for the state variable later on in this article} and taking the softmax, we can write the softmax part of the attention $A \in \mathbb{R}^{n \times n}$ in index notation as:

\begin{equation}
A_{ij} = \frac{\exp\left( \frac{q_i^T k_j}{\sqrt{d_k}} \right)}{\sum_{\text{m}} \exp\left( \frac{q_i^T k_{\text{m}}}{\sqrt{d_k}} \right)} =\frac{\exp\left( \frac{q_i^T k_j}{\sqrt{d_k}} \right)}{\zeta_i}
\label{eqn:2b}
\end{equation}

where $A \in \mathbb{R}^{n \times n}$. Also, it may be noted that $ \sum_{j=1}^n A_{ij} = 1 $ (each row of $A$ sums to 1). Including the effect of the Value matrix, $V$, we get for the attention: 

\begin{equation}
\tilde{A}= A V \in \mathbb{R}^{n \times d_v}
\label{eqn:2c}
\end{equation}

The value matrix $V$ can be written as $V = [ v_1^T v_2^T v_3^T \ldots v_n^T ]^T $ where $v_j \in \mathbb{R}^{d_v}$. Using this,  $i$th row of the attention-value product above is given as:

\begin{equation}
(A V)_i = \sum_{j=1}^n A_{ij} v_j
\end{equation}

It may be noted here that the $i$-th row of $AV)_i$ is the attention weighted sum of rows $v_j$ of the value matrix $V$.  Therfore $(AV)_1 = A_{11}v_1 + A_{12} v_2 + A_{13} v_3 + \ldots A_{1n} v_n$.  We can think of the $i$-th row $A_i$ as a focus vector, which weighs a linear combination of value vectors $v_j$ to create the attention vector for the $i$-th token\footnote{For example, if we consider the  string \texttt{``The cat killed the mouse''} with $n=5$ tokens, then the focus weights in the first row are:  $A_\text{The,The}$, $A_\text{The,cat}$, $\ldots$ $A_\text{The,mouse}$. The attention weighted first row would then be:  $A_\text{The,The} v_\text{The} + A_\text{The,cat}v_\text{cat} + \ldots A_\text{The,mouse} v_\text{mouse}$, where $v_\text{The}, v_\text{cat}, \ldots v_\text{mouse}$ are the rows of the value matrix $V$.}, $(AV)_i$.\\

We would like to propose an energy functional, whose dynamics are given by equation \ref{eqn:2} above. To motivate our discussion, consider the following construction:

\begin{equation}
E(Z) = -\text{trace}\left( Z^T \text{softmax}\left( \frac{Q K^T}{\sqrt{d_k}} \right) V \right) + \text{trace} \left( \frac{1}{2} Z^TZ \right)
\label{eqn:3}
\end{equation}

Here  $Z \in \mathbb{R}^{n \times d_v}$ is the state variable. The trace term couples $Z$ with the attention output, and its gradient drives the dynamics.  The second term is regularization term which has a quadratic structure and is hence convex.  We also note that the overall form of this function makes it convex since the first term is linear in $Z$ and does not affect the convexity. Using the definition of $A$  in equations \ref{eqn:2b} and \ref{eqn:2c} above, we can write equation \ref{eqn:3} as:

\begin{equation}
E({Z}) = - \text{trace} ({Z}^TAV) +\text{trace} (\frac{1}{2} {Z}^T{Z})
\end{equation}

To compute the trace, we utilize the Frobenius inner product, which offers a direct method to express $\text{trace}(Z^T A V)$. The Frobenius inner product for matrices $M, N \in \mathbb{R}^{n \times d_v}$ is defined as:

\begin{equation}
\langle M, N \rangle_F = \sum_{i=1}^n \sum_{k=1}^{d_v} M_{ik} N_{ik}
\end{equation}

For $Z, A V \in \mathbb{R}^{n \times d_v}$, the trace is equivalent to the Frobenius inner product, giving:

\begin{equation}
\text{trace}(Z^T A V) = \langle Z, A V \rangle_F = \sum_{i=1}^n \sum_{k=1}^{d_v} Z_{ik} (A V)_{ik}
\end{equation}

Using $(A V)_{ik} = \sum_{j=1}^n A_{ij} V_{jk}$ in the above:

\begin{equation}
E(Z) = \sum_{i=1}^n \sum_{k=1}^{d_v} Z_{ik} \sum_{j=1}^n A_{ij} V_{jk}
\end{equation}

To reorganize the summation, we interchange the order to group terms by token indices\footnote{Starting with $ \text{LHS} = \sum_{i=1}^n \sum_{k=1}^{d_v} Z_{ik} \left( \sum_{j=1}^n A_{ij} V_{jk} \right)$, move $Z_{ik}$ inside the inner sum since it has no $j$ index to give: $LHS = \sum_{i=1}^n \sum_{k=1}^{d_v} \sum_{j=1}^n Z_{ik} A_{ij} V_{jk} $. Now rearrange the order of summation: 
$\text{LHS} = \sum_{i=1}^n \sum_{j=1}^n \sum_{k=1}^{d_v} Z_{ik} A_{ij} V_{jk} $ and factor out $A_{ij}$ since it has no $k$ index,
$\text{LHS} = \sum_{i=1}^n \sum_{j=1}^n A_{ij} \left( \sum_{k=1}^{d_v} Z_{ik} V_{jk} \right) = RHS$ 
}:

\begin{equation}
E(Z) = \sum_{i=1}^n \sum_{j=1}^n A_{ij} \sum_{k=1}^{d_v} Z_{ik} V_{jk}
\end{equation}

The inner sum is the dot product between the $i$-th row of $Z$, $z_i$ and the $j$-th row of $V$, $v_j$:

\begin{equation}
 \sum_{k=1}^{d_v} Z_{ik} V_{jk} =(ZV^T)_{ij} =  z_i^T v_j
\end{equation}

where $z_i \in \mathbb{R}^{d_v}$ and $v_j \in \mathbb{R}^{d_v}$ as defined before.  Thus, the energy becomes:

\begin{equation}
E(Z) = \sum_{i=1}^n \sum_{j=1}^n A_{ij} z_i^T v_j
\label{eqn:5}
\end{equation}

or in its full form:

\begin{equation}
E(z) = -\text{trace}\left( Z^T (A V) \right) = -\sum_{i=1}^n \sum_{j=1}^n \frac{\exp\left( \frac{q_i^T k_j}{\sqrt{d_k}} \right)}{\sum_{m=1}^n \exp\left( \frac{q_i^T k_m}{\sqrt{d_k}} \right)} (z_i^T v_j)
\end{equation}

\subsection{Dynamics}

To recover the dynamics of the form given by equation \ref{eqn:2} , we write $E(Z) = -f(Z) + g(Z)$ with $f(Z) = \text{trace} ({Z}^TAV)$ and $g(Z) = \text{trace} (\frac{1}{2} {Z}^T{Z})$. The differential of $f(Z)$ is\footnote{We make use of the following identities: (i) $\text{trace}(AB) = \text{trace}(BA)$ (ii) $\text{trace}(A) = \text{trace}(A^T)$ and (iii) the total derivative of the trace of a matrix function $f(X)$ is $d(\text{trace}(f(X))) = \text{trace}(d(f(X)))$.}:

\begin{equation}
df = d(\text{trace}(Z^T AV)) = \text{trace}(d(Z^T AV)) = \text{trace}((dZ)^T AV)
\label{eqn:8}
\end{equation}  

since $AV$ is a constant matrix. We also know that for a scalar function $f(Z)$, its total derivative can be expressed in terms of its gradient with respect to $Z$ as:

\begin{equation}
df = \text{trace}\left( \left( \frac{\partial f}{\partial Z} \right)^T dZ \right)
\label{eqn:9}
\end{equation}

From Equations \ref{eqn:8} and \ref{eqn:9}, using $\text{trace}(M) = \text{trace}(M^T)$, we get:

\begin{equation}
\frac{\partial f}{\partial Z} = AV
\label{eqn:12}
\end{equation}

From the second term we get $dg =  \text{trace} \left( d(\frac{1}{2} {Z}^T{Z}) \right) = \text{trace} \left(  \frac{1}{2}(dZ^TZ + Z^TdZ) \right) 
$. But since $\text{trace} A = \text{trace} A^T$, we can say that  $dg = dZ^TZ$.  We set $Z=Z_0$ when $dE = 0$, giving the result: 

\begin{equation}
Z_0 = AV = \text{softmax}\left(\frac{QK^T}{\sqrt{d_k}}\right) 
\label{eqn:15}
\end{equation}

Thus we recover the dynamics at the minimum of the energy functional. This term appears in an energy function designed to model dynamics where the state variable \( Z \in \mathbb{R}^{n \times d_v} \) evolves toward the attention output:

\[
Z^{(k+1)} = Z^{(k)} + \eta \; \text{softmax}\left( \frac{Q K^T}{\sqrt{d_k}} \right) V
\]

and $\eta$ is the step size commonly called the Hebbian learning rate.  Since the energy is linear in $Z$ no iterations are required.  But this raises the question if we were to construct an energy functional which has a non-linear dependence on $Z$.  This is what we turn to next.

\section{Non-linear Attention via Modern Hopfield Networks}

Consider the energy functional we have derived above in equation \ref{eqn:5}:

\begin{align*}
E(Z) = \sum_{i=1}^n \sum_{j=1}^n A_{ij} z_i^T v_j
\end{align*}

and compare it to the form of the MHN energy given by equation \ref{eqn:1a} reproduced below:

\begin{align*}
E = \sum_{\mu=1}^K F\left( \sum_{i=1}^N \xi_i^\mu s_i \right)
\end{align*}

To ease the comparison, let us set:

\begin{equation}
j = \mu; \quad \xi_i^\mu = A_{i\mu} = A^\mu_i, \quad s_i = z_i, \quad K = n
\end{equation}

Plugging these into equation \ref{eqn:5} we and switching the order of summation we get:

\begin{equation}
E(Z) = \sum_{\mu=1}^n \left( \sum_{i=1}^n  A_i^\mu z_i^T v_\mu \right)
\label{eqn:6}
\end{equation}

It becomes immediately apparent that the attention $A$ is like stored memory patterns of Hopfield networks, $\xi_i^\mu$.  The interaction of the state vector $s_i$ in this case is modulated by taking the projection of $z_i$ with the value vector corresponding to the $\mu$-th pattern, $v_\mu$.  Thus the energy is a sum over token pairs, weighted by the attention scores $A_{i\mu}$, capturing the alignment between the state $z_i$ and the value vectors $v_\mu$. \\

To introduce MHN-like dynamics in equation \ref{eqn:6}, we apply a nonlinear function $F$, inspired by dense associative memories \cite{Krotov2016}:

\begin{equation}
E(Z) = \sum_{\mu=1}^n F\left( \sum_{i=1}^n A_{i\mu} z_i^T v_\mu \right) 
\label{eqn:7}
\end{equation}

The energy functional aligns with the MHN framework, where $A_{i\mu}$ acts as $\xi_i^\mu$, weighting the contribution of the pattern $v_\mu$, represented by the $\mu$-th row of $V$.  \\

Further, we will find it convenient to introduce:

\begin{equation}
u_\mu = \sum_{i=1}^n A_{i\mu} z_i^T v_\mu
\label{eqn:7b}
\end{equation} 

which allows us to write Equation \ref{eqn:7} as

\begin{equation}
E(Z) = \sum_{\mu=1}^n F\left( \sum_{i=1}^n A_{i\mu} z_i^T v_\mu \right) = \sum_{\mu=1}^n  F(u_\mu)
\label{eqn:7a}
\end{equation}

It may be noted that $F(u_\mu)$ has been set to the linear form $F(u_\mu) = u_\mu$  in equation \ref{eqn:6} above. $u_\mu$ may also be expressed in matrix form as:

\begin{equation}
u_j = \text{trace}(Z^TA\mathbf{e_j}\mathbf{e_j}^TV^T) = \text{trace}(Z^TA^jV^T) 
\label{eqn:46}
\end{equation}

where $\mathbf{e_j} \in \mathbb{R}^n$ is the unit vector with all zeros except a $1$ at the $j$-th position. The matrix $A^j = A\mathbf{e_j}\mathbf{e_j}^T$ with $A^j \in \mathbb{R}^{n \times n}$ represents a matrix with all zero elements except the $j$-th column, which is the same as the $j$-th column of $A$.\\ 

In order to recover the dynamics given by equation \ref{eqn:15}, its energy wells must converge to $Z = AV$.  As we will show in the next section, this is not in general guaranteed, and $F$ needs to be chosen carefully to satisfy this condition.  However, it is often possible to choose a suitable regularization functions $R(Z)$ such that the condition $Z =AV$ is achieved at the stationary point.\\

A nonlinear $F$ ensures iterative dynamics, converging to the attention output, as in MHNs. This interpretation frames attention as a pattern retrieval process, with $A_{ij}$ as memory coefficients, enhancing transformer training stability through energy-based regularization, building on the foundational work of Hopfield \cite{Krotov2016} and its extensions for example, Ramsauer et al. \cite{ramsauer2021hopfield}.\\

\subsection{The Motivation for Non-Linear Attention Mechanisms}

Cubic splines are widely used as a piece wise non-linear technique for capturing non linear behavior when plotting curves.  Cubic splines enable a smoother and richer visual representation of non-linear data.  Similarly, non-linear attention mechanisms have the the potential to (i) capture complex relationships and model more intricate dependencies beyond linear interactions, (ii) improve representation learning and facilitate richer and more abstract representations of queries, keys, and values and (iii) address efficiency trade-offs and potentially achieve better performance for similar computational costs or bridge the gap between linear and standard attention.\\

Some examples of earlier non-linear attention mechanisms involves employing non-linear functions to compute the similarity between queries and keys as given by Bahdanau etal. \cite{bahdanau2014neural} which utilizes a feed-forward network with a $\tanh$ activation to determine the compatibility score:
$$ \text{score}(q, k) = v_a^T \tanh(W_a[q; k]) $$

This method, initially proposed for Neural Machine Translation, allows for learning more complex relationships than a simple dot product.\\

Kernel methods afford another example, see for example the Performer architecture \cite{choromanski2020rethinking}, leverage kernel functions to redefine the attention mechanism. This introduces non-linearity through the kernel function $k$, enabling the model to capture intricate interactions between tokens:

$$ \text{Attention}(Q, K, V)_i = \sum_{j} \alpha_{ij} V_j \quad \text{where} \quad \alpha_{ij} = \frac{k(Q_i, K_j)}{\sum_{l} k(Q_i, K_l)} $$

While the Performer focuses on efficient attention using kernel methods, the underlying principle introduces non-linearity in the similarity computation.\\

Non-linear techniques are also being explored to mitigate issues like gradient instability or attention dilution. Efficient Transformer variants, such as the Nyströmformer \cite{chen2021skyformer}, while aiming for linear complexity, might incorporate non-linear elements in the aggregation process.\\

In the next few sections we consider the implications of using energy functionals of the type given by equation \ref{eqn:7a} and examine a few different forms of $F(u_\mu)$.  We begin with perhaps the simplest non-linear form of $F$, the quadratic, which also corresponds to the the form of the energy functional for classical Hopfield Networks. \\

\section{Quadratic Energy Form}

We consider a quadratic energy form given by:

\begin{equation}
F(u_j) = u_j^2 = \left( \sum_{i=1}^n A_{ij} z_i^T v_j \right)^2
\label{eqn:43}
\end{equation}

which gives us an energy functional with the familiar quadratic form\footnote{This form of the quadratic will lead to energy maxima.}:

\begin{equation}
E(Z) = \sum_{j=1}^n  \left(\sum_{i=1}^n A_{ij} z_i^T v_j \right)^2
\label{eqn:20}
\end{equation}  

To compute the gradient, we note that:

\begin{equation}
\nabla_Z E = \frac{\partial E}{\partial Z_{ik}} = \frac{\partial }{\partial Z_{ik}} \left( \sum_{j=1}^n F(u_j) \right) =   \sum_{j=1}^n  F^\prime(u_j) \frac{\partial u_j}{\partial Z_{ik}} 
\end{equation}

Now, $u_j$ as defined by equation \ref{eqn:7b} can be written as:

\begin{equation}
u_j = \sum_{m=1}^n A_{mj} \sum_{l=1}^{d_v} Z_{ml} V_{jl}
\end{equation}

Using $\frac{\partial Z_{ml}}{\partial Z_{ik}} = \delta_{im}\delta_{kl}$, the derivative may be written as:

\begin{equation}
\frac{\partial u_j}{\partial Z_{ik}} = \sum_{m=1}^n A_{mj} \sum_{l=1}^{d_v} V_{jl} \delta_{mi} \delta_{lk} = A_{ij} V_{jk}
\end{equation}

Thus the gradient is given by:

\begin{equation}
\frac{\partial E}{\partial Z_{ik}} = \sum_{j=1}^n F'(u_j) A_{ij} V_{jk}
\label{eqn:40}
\end{equation}

For the quadratic energy form given by equation \ref{eqn:43}, $F^\prime(u_j) = 2 u_j = 2 \sum_{m=1}^n A_{mj} z_i^Tv_j$.  This gives for the energy gradient:

\begin{equation}
(\nabla_Z E)_{ik} =\frac{\partial E}{\partial Z_{ik}} = \sum_{j=1}^n 2 \left( \sum_{m=1}^n A_{mj} z_m^T v_j \right) A_{ij} V_{jk}
\label{eqn:45}
\end{equation}

\subsection{Finding  $ \nabla_Z E $ at $ Z = A V $}

In order to match the dynamics given by equation \ref{eqn:15}, we need to ensure that $Z=AV$ is satisfied at the minimum of the energy functional defined by equation \ref{eqn:20}.  To evaluate the $\nabla_Z E$ under $Z=AV$, first it may be recalled that:

\begin{align*}
z_i = \sum_{p=1}^n A_{ip} v_p, \quad Z_{il} = \sum_{p=1}^n A_{ip} V_{pl}.
\end{align*}

$z_m^T v_j$ may then be evaluated under $Z=AV$ as:

\begin{equation}
z_m^T v_j = \sum_{l=1}^{d_v} Z_{ml} V_{jl} = \sum_{l=1}^{d_v} \left( \sum_{p=1}^n A_{mp} V_{pl} \right) V_{jl} = \sum_{p=1}^n A_{mp} \sum_{l=1}^{d_v} V_{pl} V_{jl} = \sum_{p=1}^n A_{mp} v_p^T v_j,
\label{eqn:44}
\end{equation}

Plugging equation \ref{eqn:44} into equation \ref{eqn:45}, we get:

\begin{equation}
(\nabla_Z E)_{ik} = \sum_{j=1}^n 2 \left( \sum_{m=1}^n A_{mj} \sum_{p=1}^n A_{mp} v_p^T v_j \right) A_{ij} V_{jk}.
\label{eqn:60}
\end{equation}

Rewrite the inner term:

\begin{equation}
\sum_{m=1}^n A_{mj} \sum_{p=1}^n A_{mp} v_p^T v_j = \sum_{m=1}^n \sum_{p=1}^n A_{mj} A_{mp} v_p^T v_j = v_j^TA^TAv_j = ||Av_j||^2 
\label{eqn:50} 
\end{equation}

where $||Av_j||^2$ is the $L_2$ norm of $Av_j$.  We set:

\begin{equation}
c_j = ||Av_j||^2 
\end{equation}

which allows us to write the above equation \ref{eqn:60} as:

\begin{equation}
(\nabla_Z E)_{ik} = \sum_{j=1}^n 2 c_j  A_{ij} V_{jk}.
\label{eqn:31}
\end{equation}

The above equation gives the gradient matrix  at $Z = AV$.  Defining a diagonal matrix $C \in \mathbb{R}^{n \times n}$ given by $C = \text{diag}(c_j)$, i.e., that the $c_j$ form the diagonal elements of $C$, we can write equation \ref{eqn:31} as:

\begin{equation}
\nabla_Z E = 2\;ACV
\end{equation} 

and it is clear that $\nabla_Z E|_{Z=AV} \neq 0$.  However, it is always possible to ensure that the gradient is zero by adding a suitable regularization term to the energy functional to create a new ``regularized'' energy functional $E_R(Z)$  as follows:

\begin{equation}
E_R(Z)  = E(Z) + R(Z)
\end{equation}

We require a suitable regularization term $ R(Z) $ such that:

\begin{equation}
\nabla_Z E_{\text{R}} = \nabla_Z E + \nabla_Z R = 0 \text{ at } Z = A V
\label{eqn:32}
\end{equation}

\subsection{Regularization}

From equation \ref{eqn:32}, it is clear that  at $Z=AV$, the regularization function needs to satisfy the condition:

\begin{equation}
\nabla_Z R(Z) = -\nabla_Z E = -2\;ACV
\end{equation}

The simplest function that satisfies this is:

\begin{equation}
R(Z) = -2\;\text{trace}(Z^TACV) 
\end{equation}

Therefore we can write for the regularized energy, $E_R$:

\begin{equation}
E_R(Z) = \sum_{j=1}^n[\text{trace}(Z^TA^jV^T)]^2 - 2\;\text{trace}(Z^TACV)
\end{equation}

which can also be written in the form:

\begin{equation}
E_R(Z) = E(Z) + R(Z) = \sum_{j=1}^n \left( \sum_{i=1}^n A_{ij} z_i^T v_j \right)^2 - 2 \sum_{i=1}^n \sum_{j=1}^n c_j A_{ij} z_i^T v_j.
\end{equation}

One important observation is that with \( F(u_j) = u_j^2 \), \( \xi_i^j = A_{ij} \), \( s_i = z_i \), the transformer energy resembles a positive quadratic form, requiring regularization to recover transformer attention outputs, unlike Hopfield networks where attractors are at pattern states.\\

In the next section we will examine a polynomial form of $F$.  

\section{Polynomial Energy Form}

Now we generalize the results of the previous section to a general polynomial form of $F(u_j)$ given by:

\begin{equation}
F(u_j) = \left( \sum_{m=1}^n A_{mj} z_m^T v_j \right)^p
\end{equation}

where $p \geq 1$ and

\begin{equation}
u_j = \sum_{m=1}^n A_{mj} z_m^T v_j
\end{equation}

The energy $E(Z)$ is then given by:

\begin{equation}
E(Z) = \sum_{j=1}^n F(u_j) = \sum_{j=1}^n \left( \sum_{m=1}^n A_{mj} z_m^T v_j \right)^p
\end{equation}

Using the notation of equation \ref{eqn:46}, we can write this as:

\begin{equation}
E(Z) = \sum_{j=1}^n [\text{trace}(Z^TA^jV^T)]^{p}
\end{equation}

To enforce a stationary point at $ Z = A V $ for the energy functional and we may need to design a suitable regularizayion function $R(Z)$ to enfore the condition.  The gradient is:

\begin{equation}
\nabla_Z E = \frac{\partial E}{\partial Z_{ik}} = \sum_{j=1}^n \frac{\partial F(u_j)}{\partial Z_{ik}}
\end{equation}

Compute the derivative with respect to \( u_j \):

\begin{equation}
F^\prime(u_j) = p u_j^{p-1} = p \left( \sum_{m=1}^n A_{mj} z_m^T v_j \right)^{p-1}
\end{equation}

which allows us to write $\nabla_Z E$ as:

\begin{equation}
\frac{\partial E}{\partial Z_{ik}} = \sum_{j=1}^n p \left( \sum_{m=1}^n A_{mj} z_m^T v_j \right)^{p-1} A_{ij} V_{jk}
\end{equation}

At \( Z = A V \):

\begin{equation}
(\nabla_Z E )_{ik} = \left( \frac{\partial E}{\partial Z_{ik}} \right) = \sum_{j=1}^n p \left( \sum_{m=1}^n A_{mj} \sum_{l=1}^n A_{ml} v_l^T v_j \right)^{p-1} A_{ij} V_{jk}
\end{equation}

Using equation \ref{} for the term in the parenthesis, we get:

\begin{equation}
(\nabla_Z E )_{ik} =  \sum_{j=1}^n p\;c_j^{p-1} A_{ij} V_{jk} = AC^{p-1}V
\end{equation}

Therefore, we need to find an $R(Z)$ such that:

\begin{equation}
\nabla_Z R(Z) = -AC^{p-1}V
\end{equation}

Therefore propose a form $R(Z)$ as follows:

\begin{equation}
R(Z) = -p\;\;\text{trace}(Z^TAC^{p-1}V)
\end{equation}

We can therefor define the regularized energy which enforces $Z=AV$ as:

\begin{equation}
E_R(Z) = \sum_{j=1}^n [\text{trace}(Z^TA^jV^T)]^{p} -p\;\;\text{trace}(Z^TAC^{p-1}V) 
\end{equation}

or in index notation:

\begin{equation}
E_R(Z) = \sum_{j=1}^n F(u_j)  =\sum_{j=1}^n \left( \sum_{m=1}^n A_{mj} z_m^T v_j \right)^p -p \sum_{i=1}^n \sum_{j=1}^n \left( \sum_{m=1}^n A_{mj} \sum_{l=1}^n A_{ml} v_l^T v_j \right)^{p-1} A_{ij} z_i^T v_j
\end{equation}

The general polynomial \( F(u_j) \) with the proposed regularization ensures a stationary point at \( Z = A V \), aligning with the Vaswani attention mechanism. For \( p \geq 2 \), the energy suggests a minimum.

\section{Exponential Energy Forms}

We now consider an exponential energy form given by:

\begin{equation}
F(u_j) = e^{u_j}= \exp \left( \sum_{m=1}^n A_{mj} z_m^T v_j \right)
\label{eqn:46}
\end{equation}

The energy would then be given by:

\begin{equation}
\sum_{j=1}^n e^{u_j}
\end{equation}

For the exponential form given by equation \ref{eqn:46}, $F^\prime(u_j) = F(u_j)$.  Therefore, using equation \ref{eqn:40}, the gradient is given by:

\begin{equation}
(\nabla_Z E)_{ik} = \frac{\partial E}{\partial Z_{ik}} = \sum_{j=1}^n \exp \left( \sum_{m=1}^n A_{mj} z_m^T v_j \right) A_{ij} V_{jk}
\label{eqn:51}
\end{equation}

At \( Z = A V \), again using the form of $z_m^T v_j$ given by equation \ref{eqn:44} in equation \ref{eqn:51} and then reorganizing the terms as in equation \ref{eqn:50}, and using $c_j = ||Av_j||^2$, we can write the above equation \ref{eqn:51} as:

\begin{equation}
(\nabla_Z E)_{ik} = \sum_{j=1}^n e^{c_j} A_{ij} V_{jk}
\end{equation}

Using the matrix $C \in \mathbb{R}^{n \times n}$ and given as $C =\text{diag}(e^{c_j})$, the above can be written as:

\begin{equation}
\nabla_Z E = ACV
\label{eqn:52}
\end{equation}

Using equation \ref{eqn:52} above, we can now define the regularization energy as:

\begin{equation}
R(Z) = Z^TACV
\end{equation}

which can be written using index notation as:

\begin{equation}
R(Z) = -\sum_{i=1}^n \sum_{j=1}^n \exp \left( \sum_{m=1}^n A_{mj} \sum_{l=1}^n A_{ml} v_l^T v_j \right) A_{ij} z_i^T v_j
\end{equation}

The regularized energy, $E_R$, can be written as:

\begin{equation}
E_R(Z) = E(Z) + R(Z)
\end{equation}

\section{An Algorithm for Non-Linear Attention Heads}

Below we develop an algorithm showing how the non-linear attention mechanisms can be incorporated into (say) BERT-like models, see Devlin et al. \cite{devlin2019bert}.  It may be noted that these techniques can also be applied for image, audio and video based transformer architectures.\\ 

First we define the token matrix, $X \in \mathbb{R}^{n \times d}$, and the weight matrices $W_q, W_k \in \mathbb{}^{d \times d_k}$ and the weight matrix for the value matrix $W_v \in \mathbb{}^{d \times d_v}$ where $d$ is the dimension of the input token embeddings.  The query, key and value matrices, $Q,K, V$ already introduced previously, are then defined in terms of the weight matrices and the token matrix $X$ as follows:

\begin{equation}
Q = X W_q, \quad K = X W_k, \quad V = X W_v,
\end{equation}

Based on the matrices $Q,K,V$, the matrix $A$ can be defined as follows:

\begin{equation}
S_{mj} = \frac{q_m^T k_j}{\sqrt{d_k}}, \quad A_{mj} = \frac{\exp(S_{mj})}{\sum_{j=1}^n \exp(S_{mj})}.
\end{equation}

The next step is to initialize the state variable $Z$.  The initial state matrix $Z^{(0)}$ is defined as:

\begin{equation}
Z^{(0)} = A V,
\end{equation}

where \( Z^{(0)} \in \mathbb{R}^{n \times d_v} \), \( A \in \mathbb{R}^{n \times n} \), \( V \in \mathbb{R}^{n \times d_v} \).\\

Next we compute the alignment scores $u_j$ for $j = 1, \ldots, n $:

\begin{equation}
u_j = \sum_{m=1}^n A_{mj} z_m^T v_j = \sum_{m=1}^n \sum_{k=1}^{d_v} A_{mj} Z_{mk} V_{jk}.
\end{equation}

It may be noted that we can also write $u_j = \text{trace}(Z^{(0)T}AV^T)$ based on equation \ref{}.  This allows the calculation of the energy $E$:

\begin{equation}
E(Z) = \sum_{j=1}^n F(u_j).
\end{equation}

The regularization coefficients $c_j$ can be calculated as:

\begin{equation}
c_j = \sum_{m=1}^n A_{mj} \sum_{l=1}^n A_{ml} v_l^T v_j = v_j^T A^T A v_j,
\end{equation}

The regularization term can be calculated as:

\begin{equation}
R(Z) = -2 \sum_{i=1}^n \sum_{j=1}^n c_j A_{ij} z_i^T v_j = -2 \sum_{i=1}^n \sum_{j=1}^n \sum_{k=1}^{d_v} c_j A_{ij} Z_{ik} V_{jk}.
\end{equation}

The regularized energy is:

\begin{equation}
E_{\text{R}}(Z) = \sum_{j=1}^n F(u_j) - 2 \sum_{i=1}^n \sum_{j=1}^n c_j A_{ij} z_i^T v_j.
\end{equation}

To compute the gradient with respect to $Z_{ik}$:

\begin{equation}
(\nabla_Z E)_{ik} = \frac{\partial E_{\text{R}}}{\partial Z_{ik}} = \frac{\partial E}{\partial Z_{ik}} + \frac{\partial R}{\partial Z_{ik}} = \sum_{j=1}^n F'(u_j) A_{ij} V_{jk} - 2 \sum_{j=1}^n c_j A_{ij} V_{jk} = \sum_{j=1}^n [F'(u_j) - 2 c_j] A_{ij} V_{jk}.
\end{equation}

The gradient matrix is:

\begin{equation}
\nabla_Z E_{\text{R}} = A [f'(u) - 2 c] V^T,
\end{equation}

where \( f'(u) = [F'(u_1), \ldots, F'(u_n)]^T \), \( c = [c_1, \ldots, c_n]^T \).

The gradient descent for \( t = 0, \ldots, T-1 \):

\begin{equation}
Z_{ik}^{(t+1)} = Z_{ik}^{(t)} - \eta \sum_{j=1}^n [F'(u_j^{(t)}) - 2 c_j] A_{ij} V_{jk},
\end{equation}

or in matrix form:

\begin{equation}
Z^{(t+1)} = Z^{(t)} - \eta A [f'(u^{(t)}) - 2 c] V^T.
\end{equation}

At $j=T$, $Z^{(T)}$ is computed and returned.\\

Compute loss gradients with respect to \( Z, Q, K, V \) for backpropagation through the network. Update the weight matrices $W_q,W_k,W_v$ via backpropagation.   \\

The algorithm integrates with BERT’s (see \cite{devlin2019bert}) multi-head attention by applying non-linear optimization per head. It enhances MLM by capturing non-linear dependencies, with added cost \( O(n^3 + T n^2 d_v) \) manageable via GPU parallelization. Numerical stability is ensured through gradient clipping.\\

\subsection{Comparison of the Computation of Linear and Non-Linear Heads}

The algorithm presented above for $F(u_j) = u_j$ (linear) and $F(u_j) = u_j^2$ (non-linear) case is presented in Table 1 to show clearly the differences.  It should be clear that for the linear case, there is no need for any steepest descent and hence steps 4, 5, 6 and 7 are not required.\\

The cost of the steepest descent for the non-linear case is \( O(n^3 + T n^2 d_v) \) as mentioned above.\\

\newpage

\begin{table}[H]
\centering
\small
\begin{tabular}{|c|p{4cm}|p{4cm}|p{4cm}|}
\hline
\textbf{Step} & \textbf{\( F(u_j) = u_j \)} (Linear) & \textbf{\( F(u_j) = u_j^2 \)} (Quadratic) & \textbf{Comments on Differences} \\
\hline
\multirow{3}{*}{\textbf{1: Attention Weights}} & Compute \( Q = X W_q \), key \( K = X W_k \), value \( V = X W_v \). & Same as linear. & No difference; attention weights are independent of \( F(u_j) \), adding 0 operations. \\
& Compute scores: \( S_{mj} = \frac{q_m^T k_j}{\sqrt{d_k}} \). & & \\
& Softmax: \( A_{mj} = \frac{\exp(S_{mj})}{\sum_{j=1}^n \exp(S_{mj})} \). & & \\
\hline
\textbf{2: Initialize \( Z \)} & Initialize \( Z \in \mathbb{R}^{n \times d_v} \), e.g., \( Z = A V \). & Same as linear. & No difference; initialization is agnostic to \( F(u_j) \), adding 0 operations. \\
\hline
\textbf{3: Compute \( u_j \)} & For \( j = 1, \ldots, n \): \( u_j = \sum_{m=1}^n A_{mj} z_m^T v_j \). & Same as linear. & No difference; \( u_j \) definition is identical, adding 0 operations. \\
\hline
\textbf{4: Compute \( E(Z) \)} & \( E(Z) = \sum_{j=1}^n u_j \). & \( E(Z) = \sum_{j=1}^n u_j^2 \). & Quadratic case squares \( u_j \), emphasizing large alignments and creating a non-linear energy landscape, adding \( n \) operations. \\
\hline
\multirow{3}{*}{\textbf{5: Compute \( R(Z) \)}} & \( c_j = \sum_{m=1}^n A_{mj} \sum_{l=1}^n A_{ml} v_l^T v_j \). & \( c_j = \sum_{m=1}^n A_{mj} \sum_{l=1}^n A_{ml} v_l^T v_j \). & Quadratic regularization includes \( c_j \) and a factor of 2, reflecting the non-linear gradient, adding \( n^3 + n^2 d_v \) operations for computing \( c_j \). \\
& \( R(Z) = -\sum_{i=1}^n \sum_{j=1}^n A_{ij} z_i^T v_j \). & \( R(Z) = -2 \sum_{i=1}^n \sum_{j=1}^n c_j A_{ij} z_i^T v_j \). & \\
& Compute: \( R(Z) = -\sum_{i=1}^n z_i^T \sum_{j=1}^n A_{ij} v_j \). & Compute: \( R(Z) = -2 \sum_{i=1}^n z_i^T \sum_{j=1}^n c_j A_{ij} v_j \). & \\
\hline
\multirow{3}{*}{\textbf{6: Compute Gradient}} & \( \frac{\partial E}{\partial z_i} = \sum_{j=1}^n A_{ij} v_j \). & \( \frac{\partial E}{\partial z_i} = \sum_{j=1}^n 2 u_j A_{ij} v_j \). & Quadratic gradient depends on \( u_j \), requiring iterative updates, unlike the constant linear gradient, adding \( n^2 \) operations for extra multiplications. \\
& \( \frac{\partial R}{\partial z_i} = -\sum_{j=1}^n A_{ij} v_j \). & \( \frac{\partial R}{\partial z_i} = -2 \sum_{j=1}^n c_j A_{ij} v_j \). & \\
& \( \frac{\partial E_{\text{R}}}{\partial z_i} = 0 \). & \( \frac{\partial E_{\text{R}}}{\partial z_i} = \sum_{j=1}^n 2 (u_j - c_j) A_{ij} v_j \). & \\
\hline
\multirow{3}{*}{\textbf{7: Optimize \( Z \)}} & Set \( Z = A V \). & Gradient descent: & Quadratic case requires iterative optimization, adding complexity but exploring the energy landscape, adding \( (T-1) n^2 d_v \) operations for \( T \) iterations. \\
& & For \( t = 1, \ldots, T \): & \\
& & \( z_i^{(t+1)} = z_i^{(t)} - \eta \sum_{j=1}^n 2 (u_j^{(t)} - c_j) A_{ij} v_j \). & \\
\hline
\textbf{8: Output \( Z \)} & Use \( Z = A V \). & Use optimized \( Z \). & Quadratic \( Z \) may differ from \( A V \), affecting downstream layers, adding 0 operations. \\
\hline
\textbf{9: Backpropagation} & Compute loss gradients w.r.t. \( Q, K, V \), including \( E_{\text{R}} \). & Same, but with gradients from \( u_j^2 \). & Quadratic case has more complex gradients from \( u_j^2 \), increasing backpropagation cost by \( n^2 \) operations. \\
\hline
\end{tabular}
\caption{Implementation algorithms for linear and quadratic energy functions in a BERT-like (see \cite{devlin2019bert}) transformer, with added calculations integrated into comments.}
\end{table}

\section{Conclusion}

Non-linear functions \( F(u_j) \), such as polynomials (\( F(u_j) = u_j^p \)) or exponentials (\( F(u_j) = \exp u_j \)), enhance representation learning by capturing complex, non-linear dependencies in text. For example, a quadratic polynomial amplifies larger alignment scores, emphasizing tokens with strong contextual relevance, crucial for tasks like masked language modeling (MLM) and next sentence prediction (NSP). This enables BERT-like models (see \cite{devlin2019bert}) to better analyze intricate patterns, such as syntactic structures or semantic relationships, improving embeddings for tasks like question answering.\\

Moreover, non-linear \( F(u_j) \) improves the optimization landscape, creating a smoother, more stable surface for gradient-based training, which is vital for BERT-like models pre-training on large datasets. It also enhances robustness to noise by suppressing irrelevant tokens, critical for handling real-world text with typos or artifacts, thus boosting performance in practical NLP applications.\\

This approach integrates with energy-based learning, supporting techniques like contrastive learning to refine BERT’s discriminative abilities. Theoretically, non-linear \( F(u_j) \) connects to associative memory, aiding token representation retrieval for tasks with long-range dependencies, such as coreference resolution, providing a robust foundation for BERT’s architecture.\\

While focused on BERT-like models for text, these benefits extend to transformer architectures in audio, video, and image processing. Capturing complex dependencies and ensuring robustness to noise can enhance tasks like speech recognition, action recognition, or object detection, broadening the applicability of non-linear energy functions.\\

Implementing non-linear \( F(u_j) \) introduces challenges like computational complexity and numerical stability, especially with high-degree polynomials or exponentials. These can be addressed through efficient approximations or stabilization techniques, requiring careful hyperparameter tuning to balance expressivity and training stability.\\

In summary, non-linear energy functions enhance BERT-like models by improving representation learning, optimization, robustness, and theoretical grounding. These advantages can elevate BERT’s performance across NLP tasks, with future work needed to empirically validate and efficiently implement this approach.

\newpage
\bibliography{Bib}
\bibliographystyle{amsplain}

\end{document}